%% file: root.tex
\pdfoutput=1

\documentclass[letterpaper, 10pt, conference]{ieeeconf}      
 
\IEEEoverridecommandlockouts                              

\usepackage{algorithm}
\usepackage{multicol}
\usepackage{multirow}
\usepackage{algpseudocode}
\usepackage{color}
\usepackage{xcolor}
\usepackage{graphicx, epstopdf}
\usepackage{bm}
\usepackage{amsmath}
\usepackage{amssymb}
\usepackage{tabularx}
\usepackage{subcaption}
\usepackage{url}
\usepackage{cite}
\usepackage{prettyref}
\usepackage{booktabs}
\usepackage{siunitx}

\newrefformat{fig}{Figure~\ref{#1}}
\newrefformat{sec}{Section~\ref{#1}}
\newrefformat{tab}{Table~\ref{#1}}
\newrefformat{alg}{Algorithm~\ref{#1}}
\newrefformat{eqn}{Equation~\ref{#1}}

\usepackage{hyperref}
\hypersetup{
    colorlinks,
    linkcolor={blue!80!black},
    citecolor={blue!80!black},
    urlcolor={blue!100!black}
}

\makeatletter
\newcommand{\printfnsymbol}[1]{%
  \textsuperscript{\@fnsymbol{#1}}%
}
\makeatother

\title{
\LARGE \bf 
Learning Resilient Behaviors for Navigation Under Uncertainty
} 

\author{Tingxiang Fan, Pinxin Long, Wenxi Liu, Jia Pan$^\dagger$, Ruigang Yang, and Dinesh Manocha%
\thanks{T. Fan, J. Pan are with the University of Hong Kong. P. Long and R. Yang are with the Robotics and Auto-Driving Lab, Baidu Research. W. Liu is with Fuzhou University. 
D. Manocha is with the University of Maryland, College Park.  $^\dagger$ denotes the corresponding author. 
}
\thanks{This work was partially supported by HKSAR General Research  Fund (GRF) HKU 11202119, 11207818, and NSFC/RGC Joint Research Scheme (HKU103/16-NSFC61631166002)}
}

\begin{document}
\maketitle

\begin{abstract}

Deep reinforcement learning has great potential to acquire complex, adaptive behaviors for autonomous agents automatically. However, the underlying neural network polices have not been widely deployed in real-world applications, especially in these safety-critical tasks (e.g., autonomous driving). One of the reasons is that the learned policy cannot perform flexible and resilient behaviors as traditional methods to adapt to diverse environments. In this paper, we consider the problem that a mobile robot learns adaptive and resilient behaviors for navigating in unseen uncertain environments while avoiding collisions. We present a novel approach for uncertainty-aware navigation by introducing an uncertainty-aware predictor to model the environmental uncertainty, and we propose a novel uncertainty-aware navigation network to learn resilient behaviors in the prior unknown environments. To train the proposed uncertainty-aware network more stably and efficiently, we present the temperature decay training paradigm, which balances exploration and exploitation during the training process. Our experimental evaluation demonstrates that our approach can learn resilient behaviors in diverse environments and generate adaptive trajectories according to environmental uncertainties. 



\end{abstract}

\input{intro}

\input{background}

\input{problem}

\input{approach}

\input{results}

\input{conclusion}

\clearpage
{\small
\bibliographystyle{IEEEtran}
\bibliography{references}
}

\end{document}

%% file: intro.tex
\section{Introduction}
\label{sec:intro}

With the recent progress of machine learning techniques, deep reinforcement learning has been seen as a promising technique for autonomous systems to learn intelligent and complex behaviors in manipulation and motion planning tasks~\cite{levine2018learning, bojarski2016end, fan2019getting}. However, due to the difficulty of interpreting and manipulating deep neural networks (DNNs), they are often not as flexible as traditional methods \cite{fox1997dynamic} that can adapt to diverse environments by parameter-tuning. Furthermore, the performance of DNNs may significantly decrease when the test data distribution is very different from the training data. Thus, DNNs may cause catastrophic failures in safety-critical tasks such as robot navigating in real-world human crowds \cite{kahn2017uncertainty}. 

To address the above issues, some related work~\cite{tobin2017domain, long2018towards, peng2018sim} attempted to expand the domain of the training data by increasing its diversity. 
However, it is impossible to completely cover all testing domains and all scenarios that robots may encounter. For example, although there are many synthetic crowd simulation techniques based on sim-to-real paradigm ~\cite{narain2009aggregate, van2011reciprocal, curtis2016menge}, it is still difficult to simulate realistic pedestrian behaviors that would take into account the underlying human-robot interaction. 
In contrast, humans are able to adaptively adopt behaviors to interact with unseen behaviors and scenes according to their sensed uncertainty about the surroundings. For example, a car-driver will slow down to maintain a safe distance from other vehicles when driving in snow. Our work aims to develop a human-alike uncertainty-aware navigation policy for robots to handle unseen environments robustly and efficiently. 

Recently, several methods are proposed in the deep learning community on uncertainty estimation, which has been widely applied to predict the model outcome in perception and inference tasks~\cite{gal2016uncertainty, kendall2017uncertainties, wang2018end}. Many techniques have accomplished this by revising the architecture of DNNs and their loss functions. In particular, the aleatoric uncertainty can be modeled by a specific loss function for the uncertainty term in the network output and the epistemic uncertainty (i.e. model uncertainty) can be captured by the Monte Carlo Dropout (MC-Dropout) technique, which randomly samples from the network via dropout layers at test time ~\cite{kendall2018learning}. By utilizing these techniques, the measures of uncertainty have been extended to decision-making networks for imitation learning paradigms~\cite{choi2018uncertainty, tai2019visual,segu2019general}. However, in practice, it is usually expensive to collect a large number of high-quality demonstrations, especially edge cases in applications like autonomous driving or robots navigating in challenging indoor scenes.
Moreover, uncertainty estimation has also been employed in the reinforcement learning framework. Many works use uncertainty estimation to develop an uncertainty-aware model-based reinforcement learning that deploys the model-predictive control (MPC) pipeline to generate safer actions~\cite{kahn2017uncertainty, lotjens2019safe}. These studies demonstrate that learning-based controllers have great potential for safety-critical tasks due to their resilient behaviors under uncertain environments. However, the parameters of MPC optimization need careful tuning to achieve desired behaviors, otherwise may  get stuck in a bad local minimum~\cite{kahn2017uncertainty}.

\begin{figure*}
\centering
\includegraphics[width=.8\linewidth]{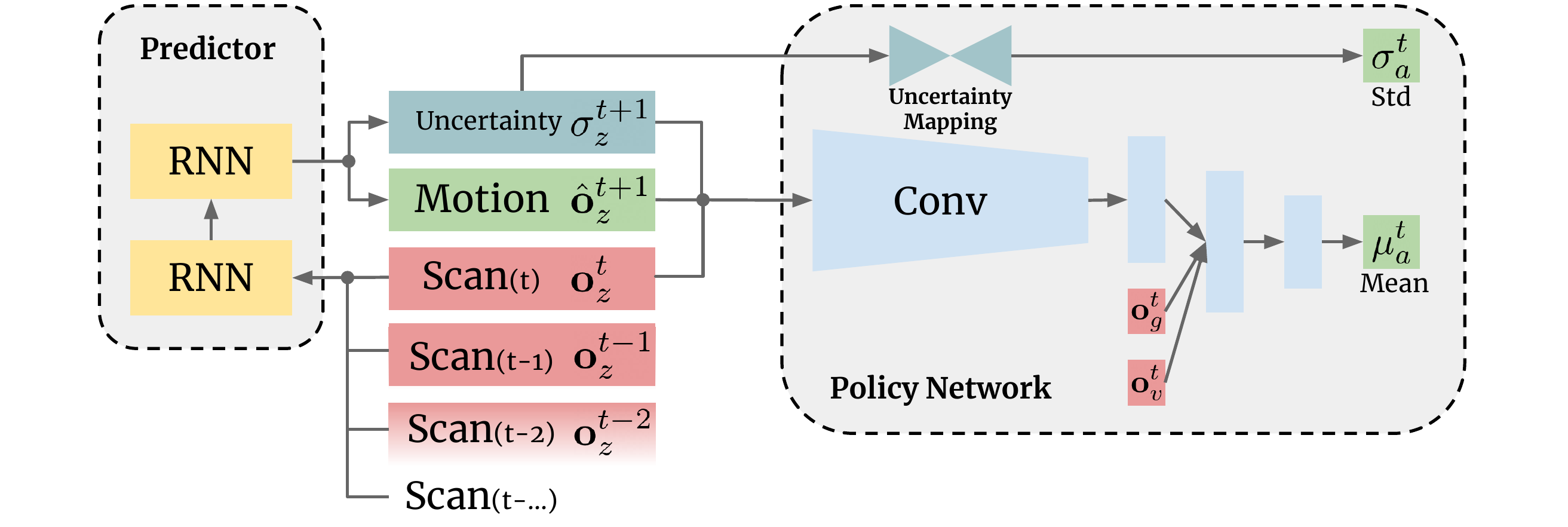}
\caption{The proposed navigation framework is mainly divided into two modules, the predictor and the policy network. Our predictor is a recurrent network that captures the uncertainty and motion information in the environment from a sequence of laser scan data. After that, the uncertainty and motion information are encoded as part of the input of the policy network to predict the action distribution. In particular, feeding the input through convolutional layers and fully-connected layers, the mean value of the action distribution can be obtained. Besides, the variance of the distribution is modeled as an uncertainty-dependent vector estimated by a uncertainty mapping function in order to enhance the connection between the uncertainty and the navigation behavior. }
\label{fig:policy_network}
\vspace*{-0.15in}
\end{figure*}

{\bf Main Contributions:} In this paper, we present techniques to show that model-free reinforcement learning can be used to learn safer and more resilient behaviors for navigation by integrating uncertainty estimation. 
With the resilient behaviors, the robot can adaptively adjust the safe distance and speed according to the uncertainty of its surrounding environment.
To learn resilient behaviors, our key idea is to bridge the gap between the environmental uncertainty and the uncertainty of the policy network's output. By doing so, environmental uncertainty can be minimized by decreasing the uncertainty of the action during the training process. Specifically, we first introduce the uncertainty-aware prediction module to capture the uncertainty and motion information of surrounding environments. We then design this paper's main contribution, an uncertainty-aware policy network with two-way connections as shown in \prettyref{fig:policy_network}, to learn resilient behaviors for navigation. Third, to train the neural network, we propose the \textit{temperature decay training paradigm} for the soft actor-critic (SAC) \cite{haarnoja2018soft} algorithm which maximizes the entropy of the policy to ensure the policy can be fully explored.

%% file: background.tex
\section{Related Work}
\label{sec:related_work}


\subsection{Uncertainty Estimation for DNNs}
The uncertainty composition in DNNs can be mainly divided into \textit{aleatoric} uncertainty and \textit{epistemic} uncertainty according to~\cite{gal2016uncertainty, kendall2017uncertainties, segu2019general}. \textit{Aleatoric} uncertainty, also known as data uncertainty, captures the uncertainty concerning observation data. \textit{Epistemic} uncertainty refers to the model uncertainty about the robot system.

To extract above two types of uncertainties, traditional Bayesian methods can be well applied to low-dimensional problems~\cite{graves2011practical, neal2012bayesian}. However, for high-dimensional and complex DNNs, more scalable and flexible approaches must be employed. Gal et al. ~\cite{gal2016dropout} present the Monte Carlo Dropout (MC-Dropout) approach to  capture the model uncertainty in DNNs by introducing dropout as a Bayesian approximation. Since the aleatoric uncertainty is related to the input data, Kendall et al.~\cite{kendall2017uncertainties} build a map from inputs to network outputs and propose a modified loss function to learn the uncertainty of the data. By combining both uncertainty estimation methods, deep learning models have achieved state-of-the-art results on computer vision tasks \cite{kendall2017uncertainties}. 

\subsection{Learning Navigation Policy with Uncertainty-Awareness}
Many techniques have been proposed that use a decision-making network for navigation tasks by imitation learning (IL)\cite{pfeiffer2017perception, long2017deep} and reinforcement learning (RL)\cite{faust2018prm,chiang2019learning}. Imitation learning trains a model by mimicking a set of desired behaviors demonstrated by experts. Based on the similarity of IL in navigation and visual recognition tasks, Tai el al.~\cite{tai2019visual} accomplish uncertainty-aware visual-based navigation by extending the use of uncertainty in visual tasks. Segu et al.~\cite{segu2019general} present a general framework that allows propagation of uncertainties in networks, providing action uncertainty and performing robust behaviors for noisy and adversarial data. Although the uncertainty-aware IL can improve the robustness and safety in autonomous driving scenarios, it is difficult to learn policies to actively avoid uncertain regions. 
 
Uncertainty estimation methods have also been applied in RL fields. Some works suggest that this kind of uncertainty can be formulated as the policy uncertainty to boost exploration~\cite{osband2016deep}. For safe-critical applications in robotics, Kahn et al.~\cite{kahn2017uncertainty} first achieve uncertainty-aware navigation in static environments by a model-based reinforcement learning method. In practice, the collision probability and uncertainty prediction are formulated into the risk term for the model-predictive controller (MPC) to minimize. Lotjens et al.~\cite{lotjens2019safe} extend~\cite{kahn2017uncertainty} to avoid dynamic obstacles in complex scenarios by proposing an ensemble of LSTM networks to estimate the uncertainty of surrounding dynamic obstacles.
 
\subsection{Model-Free Navigation Networks}
Compared to the model-based RL methods, model-free RL can learn optimal policies without relying on MPC for action selection. Chen et al. \cite{chen2017decentralized, chen2017socially} use compact agent-level information as the network input and introduce the value network to model the human-robot cooperative behaviors in dynamic environments. Nonetheless, these methods can only be applied in small-scale scenes because the resulting policy can only accept a limited amount of agent-level information as the input. Some following work ~\cite{everett2018motion, chen2018crowd} address this issue by applying a Long Short Term Memory (LSTM) network to adjust the inputs in different scales.

Unlike the navigation networks above, which unitize processed agent-level information for decisions, sensor-level networks can build direct mappings from the raw sensor data to actions. Tai et al.~\cite{tai2017virtual} train the LiDAR-based navigation policy in simulation, which can navigate a single robot in real-world static environments. Fan et al.~\cite{long2018towards,fan2018fully} propose a multi-stage, multi-scenario training framework to learn a decentralized collision avoidance policy for the multi-robot system. This sensor-level network has been successfully deployed in real-world applications, including a multi-robot system for autonomous warehouse and a service robot navigating in crowds. In this paper, we extend model-free navigation networks to learn resilient behaviors with uncertainty-awareness for the sensor-level navigation network.

%% file: problem.tex
\section{Problem Formulation}
\label{sec:problem}

The sensor-level navigation network can be formulated as a Partially Observable Markov Decision Process (POMDP) and can be solved using a reinforcement learning framework. Formally, we describe a POMDP as a 6-tuple $(\mathcal{S}, \mathcal{A}, \mathcal{T}, \mathcal{R}, \Omega, \mathcal{O})$, where $\mathcal{S}$ is a set of states ($\mathbf s \in \mathcal{S}$), $\mathcal{A}$ is a set of actions ($\mathbf a \in \mathcal{A}$), $\mathcal{T}$ is the transition probability between states $\mathcal{T}(\mathbf{s'} | \mathbf{s},\mathbf{a})$, $\mathcal{R}$ is the reward function ($\mathcal{S} \times \mathcal{A} \to \mathbb{R}$), $\Omega$ is a set of observations $\mathbf o \in \Omega$, and $\mathcal{O}$ is the observation distribution given the state $\mathbf o \sim \mathcal{O}(\mathbf s)$. In our formulation, the state at $t$ time step, $\mathbf{s}^t$, consists of the agent's position $\mathbf{p}^t$, velocity $\mathbf{v}^t$, and goal $\mathbf{g}$, i.e. $\mathbf{s}^t = \langle \mathbf{p}^t, \mathbf{v}^t, \mathbf{g} \rangle$. The action $\mathbf{a}^t$ is the steering command of a differential robot in terms of linear and angular velocities. The observation  at time $t$ is $\mathbf{o}^t$, which includes the 2D laser scan $\mathbf{o}_z^t$, the relative goal $\mathbf{o}_g^t$, and the robot's current velocity $\mathbf{o}_v^t$. That is, $\mathbf{o}^t = \langle \mathbf{o}_z^t, \mathbf{o}_g^t, \mathbf{o}_v^t\rangle$. The optimal policy $\mathbf{\pi^*}$ is defined according to the Bellman equation:
\begin{align}
&\mathbf{\pi^*} = \mathop{\arg\max}_{\mathbf{a^*}} r(\mathbf{s}^t) + \gamma \int_{\mathbf{s}^{t+1}} \mathcal{T}(\mathbf{s}^{t+1}|\mathbf{s}^t, \mathbf{a}^t)V^{*}(\mathbf{s}^{t+1})\mathrm{d}\mathbf{s}^{t+1}, \notag \\
&V(\mathbf{s}^t) = \sum_{t'=t}^{T}\gamma^{t'-t}r(\mathbf{s}^{t'}), \label{eqn:optimal_pi}
\end{align}
where $V(\mathbf{s}^t)$ is the value function used for estimating the expected reward, $r(\mathbf{s}^t)$ is the reward given $\mathbf{s}^t$, and $\gamma$ is the discount factor in reinforcement learning. In our framework, we use the reward function defined in \cite{long2018towards}, that is:
\begin{equation} 
r^t = 
\begin{cases}
  20  & \ \text{if } \| \mathbf{p}^t - \mathbf{g} \| < 0.1 \\
  -20 & \ \text{else if \text{collision}} \\
  2.5\cdot(\|\mathbf{p}^{t-1}- \mathbf{g} \| - \|\mathbf{p}^{t}- \mathbf{g} \|) & \ \text{otherwise}. \\
\end{cases}
\label{eqn:simple_reward}
\end{equation}

%% file: approach.tex
\section{Approach}
\label{sec:approach}

In this section, we present our approach that enables the navigation network to learn resilient behaviors under uncertain environments. 
First, to extract uncertainty and motion information from environments, we introduce a predictor with uncertainty estimator. Then, we propose an uncertainty-aware policy network and demonstrate the underlying principles of resilient behavior learning. Finally, to better conform to the training process of the policy network, we adopt a training decay paradigm. 

\subsection{Predictor with Uncertainty Estimation}
In our formulation, a robot can only observe the static local map by 2D LiDAR, which is sufficient for navigation in static environments. However, for highly dynamic and complex scenarios, the motion information in the environment (e.g. the dynamic obstacles) is critical for robots to generate safe and smooth actions. Thus, a prediction network that can estimate the motion information is necessary. On the other hand, we assume that \textit{if an obstacle cannot be surely predicted, which means it has a high uncertainty}. Based on this assumption, we expect to measure the environmental uncertainty by computing the confidence of the prediction results. To this end, we use a Gated Recurrent Units (GRU)-based network~\cite{cho2014gru} as shown in the left half of \prettyref{fig:policy_network}, to extract the motion and uncertainty information in environments simultaneously:
\begin{equation}
\sigma_z^{t+1}, \hat{\mathbf{o}}_z^{t+1} = \mathbf{F}_{pred}(\mathbf{o}_z^{t-T:t}, \mathbf{o}_v^{t-T:t}),
\label{eqn:rnn}
\end{equation}
where $\sigma_z^{t+1}$ represents the uncertainty in the local map, including each laser point; $\hat{\mathbf{o}}_z^{t+1}$ is the prediction of the motion information; $\mathbf{o}_z^{t-T:t}$ and $\mathbf{o}_v^{t-T:t}$ are sequences of observations of laser scans and velocity in the past $T$ time steps. The network consists of two GRU layers, and each layer has 256 hidden units. To train this prediction network, we apply the training method for data uncertainty in \cite{kendall2017uncertainties} and construct the loss function as:
\begin{equation}
\begin{aligned}
\mathcal{L}(\theta) = \frac{\| \mathbf{o}_z - \hat{\mathbf{o}}_z \|_2}{2{\sigma_z}^2} + \frac{1}{2}\log{\sigma_z}^2,
\end{aligned}
\label{eqn:uncertainty_loss}
\end{equation}

\noindent where $\theta$ is the weight of the prediction network. The loss function discourages large uncertainty while minimizing prediction errors. 

\begin{figure}
\centering
\includegraphics[width=1\linewidth]{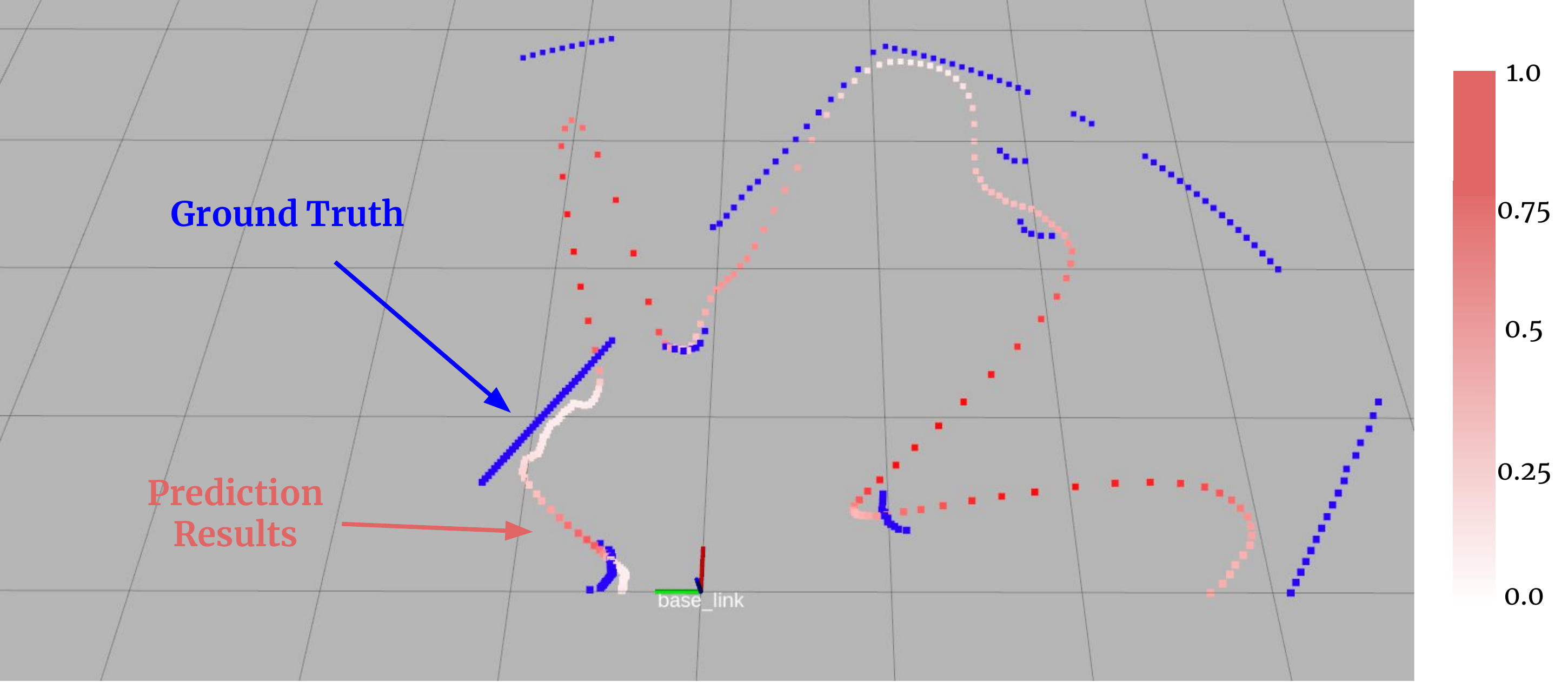}
\caption{A demonstration of prediction results: red points represent the prediction results whose brightness indicates the uncertainty value; blue points are the ground truth. }
\label{fig:pred_results}
\vspace*{-0.1in}
\end{figure}

\subsection{Uncertainty-Aware Policy Network}
\label{sec:network}

By taking advantage of uncertainty and motion information extracted by the predictor, an uncertainty-aware policy network is expected to learn resilient behaviors to maintain efficient and robust performance in diverse environments. To accomplish this, our core idea is to establish a connection from the environmental uncertainty to the uncertainty of the action distribution. Thus, we propose a uncertainty-aware policy network for model-free reinforcement learning methods. 

\begin{figure}
\centering
\includegraphics[width=0.7\linewidth]{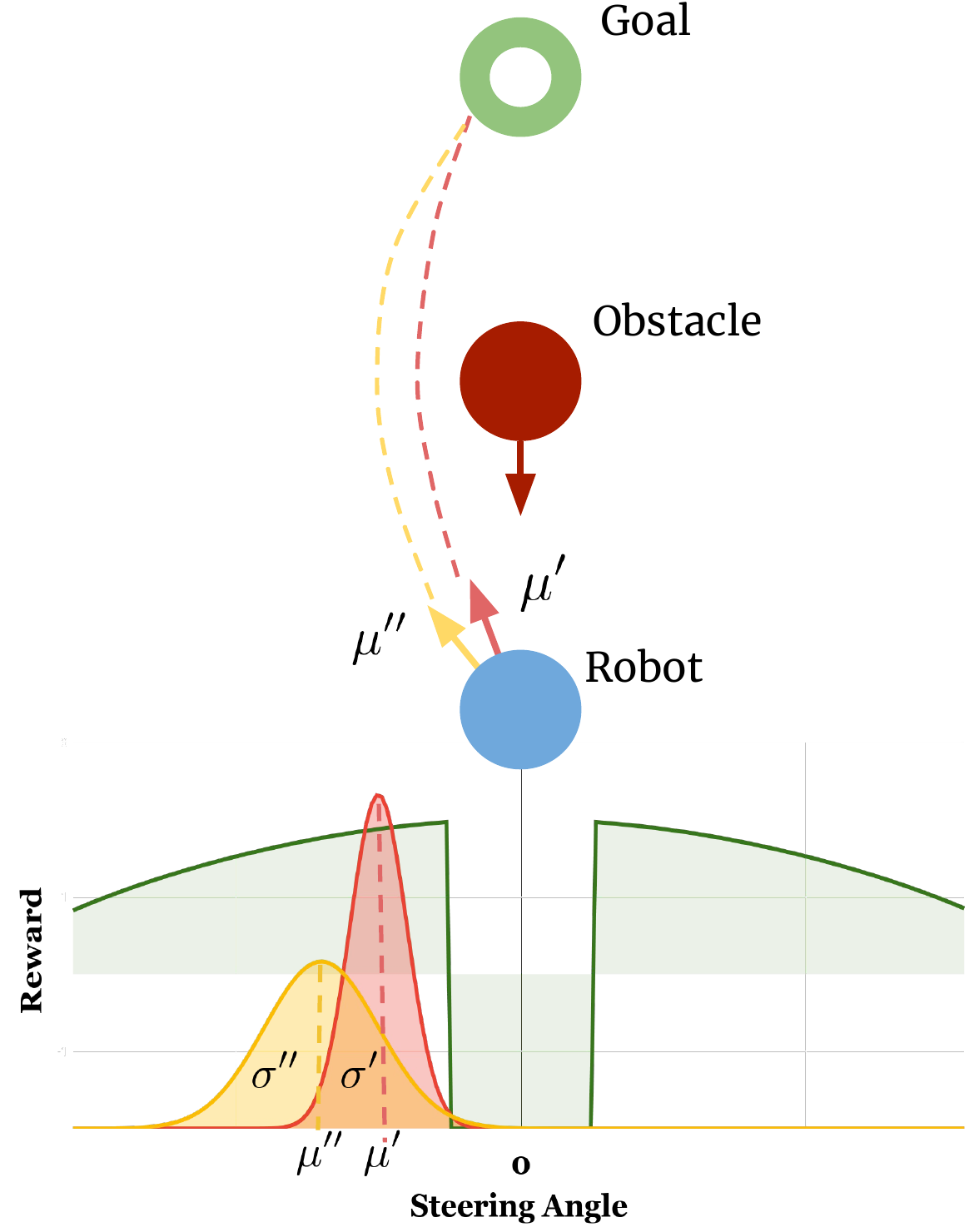}
\caption{A simple example to demonstrate the underlying idea. The uncertainty of the obstacle behavior directly affects the variance of the action distribution $\sigma$, which indirectly leads to the changes to the selection of the mean value $\mu$. Eventually, the robot can generate a more conservative trajectory for the obstacle with higher uncertainty. }
\label{fig:f2f_demo}
\vspace*{-0.15in}
\end{figure}

In contrast to previous network structure that uses multi-frame data as input to extract dynamic information in the environment \cite{mnih2015human,long2018towards,fan2018fully}, our policy network integrates the uncertainty, the motion information, and the laser scan at the current time step into the input. After feeding the rich information into the network, our formulation forwards it through two-way connections. Convolutional neural networks process one way, and the final output is the mean value of the policy distribution, which is also the learning objective of the RL algorithm. For the other way, we construct the mapping function from the environmental uncertainty to the variance of the policy distribution as follows. We first normalize the observation and environment uncertainty to $[0, 1]$ by let $\bar{\mathbf{o}} = \frac{\mathbf{o} - \mathbf{o}_{\min}}{\mathbf{o}_{\max} - \mathbf{o}_{\min}} $ and $^i\bar{\sigma}_z^t =
\frac{^i\bar{\sigma}_z^t - \sigma_{\min}}{\sigma_{\max} - \sigma_{\min}}$. 
After normalization, we consider that the closer the laser point, the higher the impact on the action, so we use the term $(2 - ^i\bar{\mathbf{o}}_z^t) \in [1, 2]$ to weight the uncertainty term $^i\bar{\sigma}_z^t$ at the $i$-th bearing accordingly: 
\begin{align}
\bar{\sigma}_{a}^{t} = \frac{\sum_{i=0}^N {} ^i\bar{\sigma}_z^t * (2 - ^i\bar{\mathbf{o}}_z^t)}{2N} \notag
\end{align}
Note that $\bar{\sigma}_a^{t}$ is also a value in the range $[0, 1]$, and thus need to re-scaled to the range of variance, resulting in the final policy variance $\sigma_a^t$:
\begin{align}
    \sigma_a^t = \sigma_{\min} + (\sigma_{\max} - \sigma_{\min}) * \bar{\sigma}_a^t, \label{eqn:uncertainty_mapping}
\end{align}
To this end, our mapping function enables the variance to reflect the uncertainty in the environment directly. Note that the variance is an uncertainty-dependent vector instead of the learnable value in the standard RL implementation. The network structure is shown in the right half of \prettyref{fig:policy_network}.

There are two benefits when using the above mentioned policy network with two-way connection. First, given different variance values, the policy network can adjust the mean value accordingly to obtain an optimal policy. Second, it allows the network to actively avert the uncertainty area during navigation (i.e. an uncertainty-averse behavior) by minimizing the entropy of the action distribution. An example of navigating a robot in environments with different uncertainties is given as follows: consider a single robot navigation task, as demonstrated in \prettyref{fig:f2f_demo}. In this task, the navigation goal (the green point) is right in front of the robot (the blue point), but they are blocked by an oncoming dynamic obstacle (the red point). To better understand this task, we roughly draw the reward distribution of the robot in the lower part. We assume that the robot prefers performing the actions on the left to avoid the obstacle and that the obstacle has two behaviors with different uncertainty levels. Thus, the environmental uncertainty will respond to the variance of the policy distribution by the mapping function  (i.e. $\sigma'$ and $\sigma''$ in the figure). To achieve the maximal reward, the navigation network can learn to adaptively generate the mean value of the distribution (i.e. $\mu'$ and $\mu''$) according to the variance value. Finally, the network performs more conservative behaviors in the case of high environmental uncertainty, that is, resilient behaviors are learned. On the other hand, by recalling the definition of the mapping function, we know the closer the environmental uncertainties, the higher the variance of the policy distribution $\sigma$. Therefore, we can obtain uncertainty-averse behaviors by minimizing the entropy of the robot trajectories.

\begin{figure}[h]
\vspace*{-0.05in}
\centering
\includegraphics[trim={30 0 15 30},clip, width=0.9\linewidth]{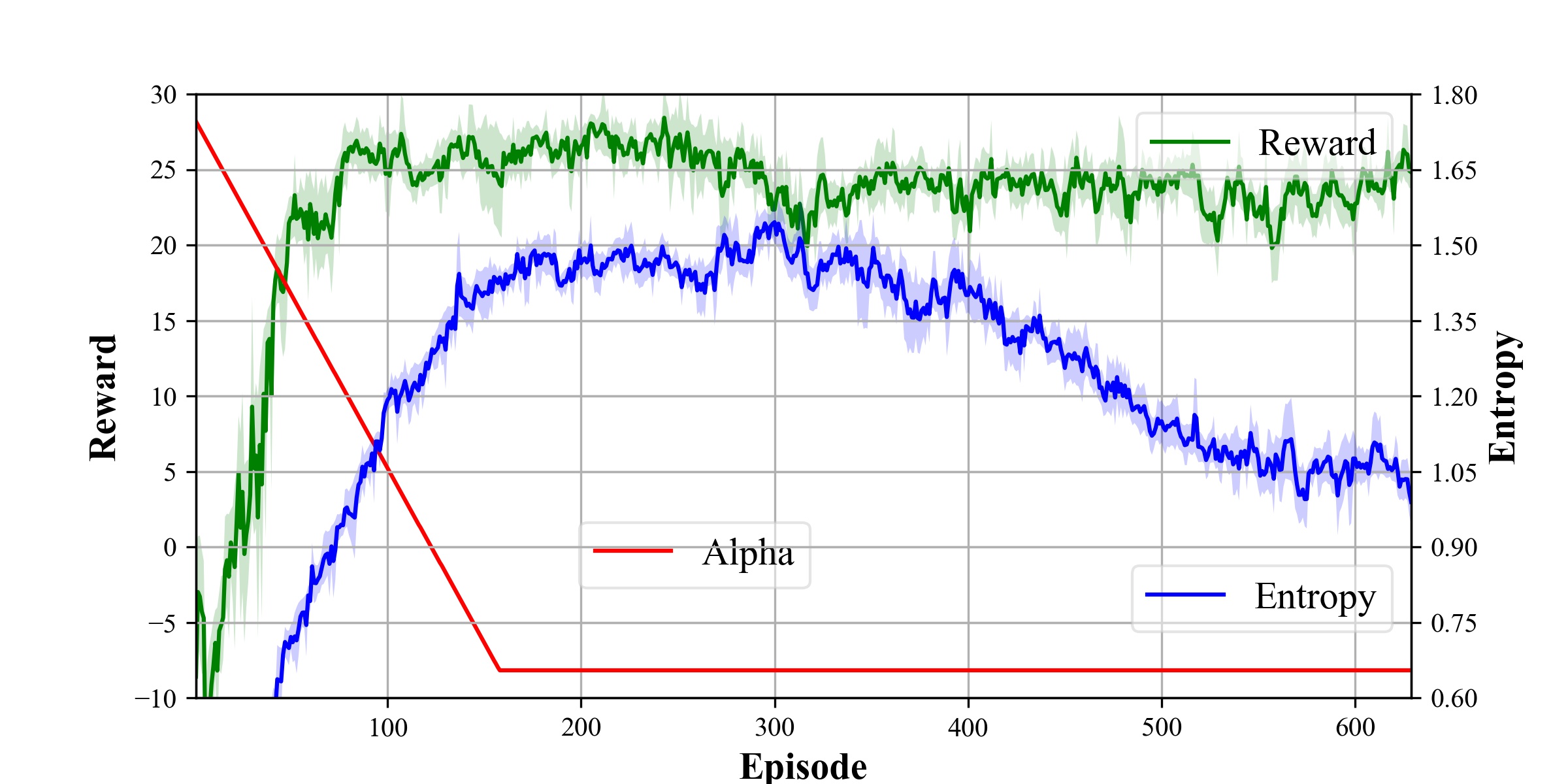}
\caption{Effect of the temperature decay paradigm on the entropy of the policy distribution and cumulative reward during the training. }
\label{fig:alpha_decay}
\vspace*{-0.05in}
\end{figure}

\subsection{Temperature Decay Training Paradigm}
\label{sec:decay}

Referring to the \textit{multi-scenario multi-stage training} framework in \cite{long2018towards}, in this paper, the training environment is based on the Stage simulator \cite{vaughan2008massively} and the multiple scenarios are designed for training. In contrast to \cite{long2018towards}, which deployed on-policy Proximal Policy Optimization (PPO) \cite{ppo} as the training algorithm, we use the Soft Actor-Critic (SAC) \cite{haarnoja2018soft} to train the uncertainty-aware navigation network. The key idea of SAC is to maximize the expected return and action entropy together instead of the expected return itself to balance the exploration and exploitation. Specifically, the origin value function defined in \prettyref{eqn:optimal_pi} is changed to include the entropy term as:
\begin{equation}
\begin{aligned}
V(\mathbf{s}^t) =  \sum_{t'=t}^{T}\gamma^{t'-t}\left(r(\mathbf{s}^{t'}) + \alpha * H(\pi(\cdot | \mathbf{s}^t))\right),
\end{aligned}
\label{eqn:soft_v}
\end{equation}
where $H(\pi(\cdot | \mathbf{s}^t))$ is the entropy of the policy distribution and $\alpha$ is the temperature for exploration-exploitation trade-off. Note that higher temperature corresponds to more exploration, and vice versa.  

There are two reasons why we replace PPO with SAC. First, the variance of policy distribution is modeled as a state-independent parameter vector in standard implementations of on-policy RL methods for stabilizing the training process. Instead, SAC parameterizes the variance as a state-dependent vector and can establish the connection between uncertainty information and variance. Second, it is challenging for PPO to train an uncertainty-aware behavior as it will discourage exploration, and the optimization may get stuck in a bad local minimum during the training process~\cite{lotjens2019safe}. SAC is better at balancing exploration and exploitation by tuning the temperature $\alpha$ during the training process. In this case, we introduce the \textit{temperature decay training paradigm} to obtain an uncertainty-aware navigation network.

The temperature decay is similar to the annealing epsilon-greedy in Deep Q-Learning Network (DQN) \cite{mnih2015human}. In the initial stage, the temperature is set a high value $\alpha_{h}$ to encourage the policy to explore the entire action space. Then, the temperature smoothly decreases by the decay rate $v_{decay}$ over the training process to focus on the exploited reward. Finally, the temperature gradually approaches the lowest bound $\alpha_l$ and stops decaying. To obtain the uncertainty-averse behavior, $\alpha_l$ is set to a negative number to encourage the lowest entropy of the policy network. 

The results of temperature decay during the training process is demonstrated in \prettyref{fig:alpha_decay}. With the increase in training episodes, the temperature decreases steadily until the $\alpha_l$. During this process, the entropy of the policy distribution (the blue line in \prettyref{fig:alpha_decay}) rises first and then decreases gradually. Note that, while the entropy is reduced, the reward of training is also decreased because the robot loses a little efficiency when it performs the uncertainty-averse behavior.

%% file: results.tex
\section{Experiments and Results}
\label{sec:exp}

\begin{figure}
\centering
\includegraphics[width=.75\linewidth]{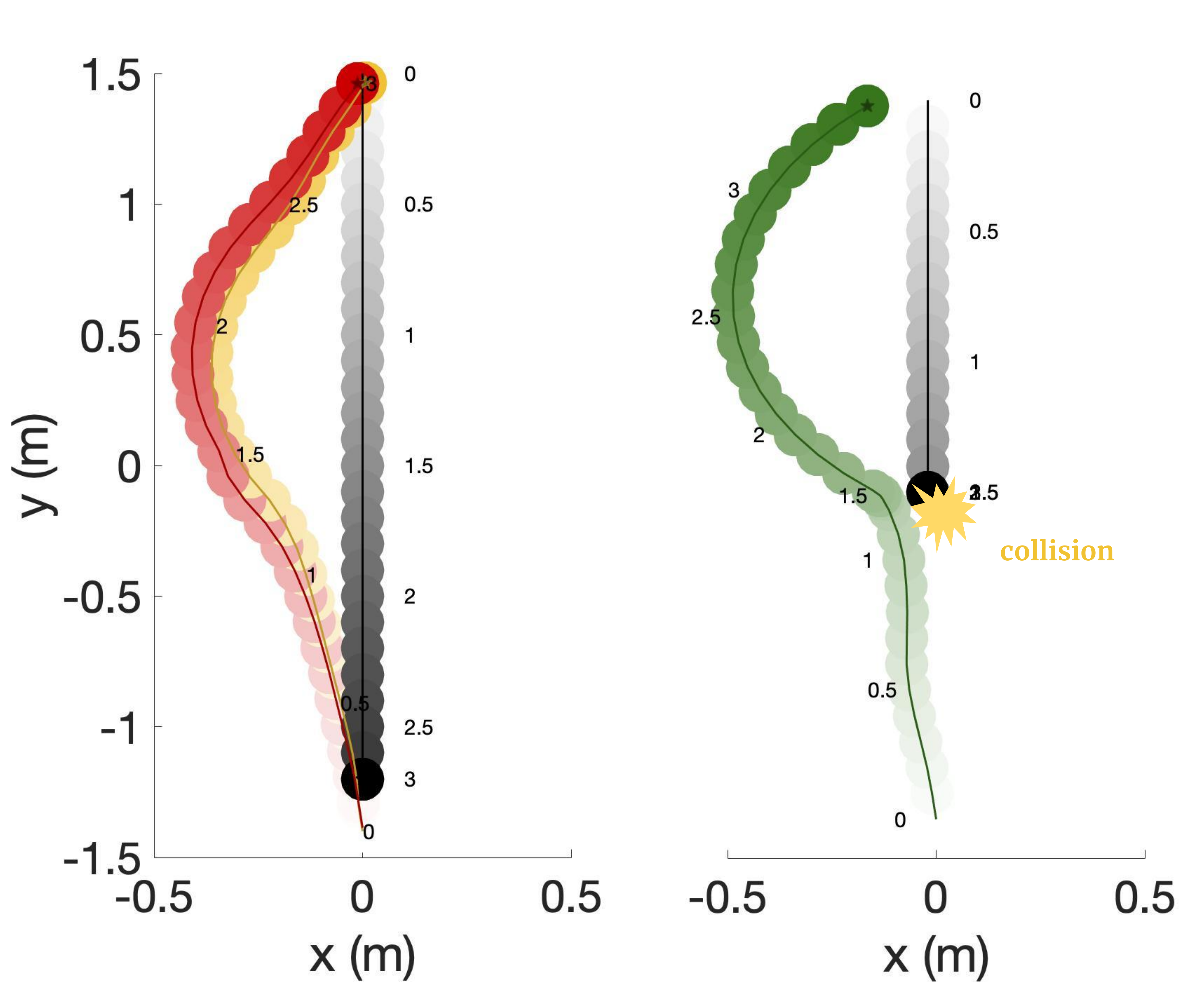}
\caption{The experimental results of different navigation policies when executing in the scenario shown in \prettyref{fig:f2f_demo}. Left: the trajectories generated by the normal (yellow) and conservative (red) policies when meeting a dynamic obstacle. Right: the overconfident policy (green) leads to a passive collision with the dynamic obstacle.}
\label{fig:f2f}
\vspace*{-0.1in}
\end{figure}


\subsection{Implementation Details}
As we mentioned above, the navigation framework mainly consists of a prediction network and a navigation network. The prediction network is trained by supervised learning, and we use the trained navigation network in \cite{long2018towards} to collect training data in the multi-scenario environment. There are about 25,000 trajectories in the dataset, $80\%$ of which are used for training; the remaining data is used for testing. We deploy the Adams optimizer with $1e^{-3}$ learning rate \cite{kingma2014adam} and $4096$ batch size in $50$ training epochs. In terms of the navigation network, the best reward can be achieved in 100 episodes. However, to obtain the uncertainty-averse behaviors, the network needs to be trained on 500 episodes, as shown in \prettyref{fig:alpha_decay}. The Adams optimizer also is applied, with a learning rate of $1e^{-3}$ before reaching 200 episodes and then with a learning rate of $1e^{-4}$ afterwards. Other hyper-parameters in this paper are summarized in \prettyref{tab:hyper}.

\begin{table}[h]
\centering
	\caption{Hyperparameters}
	\label{tab:hyper}
	\fontsize{9.5}{9.5}\selectfont
	\bgroup
	\def\arraystretch{1.3}
 \begin{tabular}{lllll} 
\hline
Parameter & Value &  & Parameter & Value  \\ 
\hline
\text{$\gamma$ in Eqn \ref{eqn:optimal_pi}}        	& $0.99$    &  & \text{$\alpha_h$ in Sec\ref{sec:decay} }        & $0.1$     \\
\text{$\sigma_{min}$ in Eqn \ref{eqn:uncertainty_mapping}}        	&  $-2.0$   &  & \text{$\alpha_l$ in Sec \ref{sec:decay}}        & $-0.01$    \\
\text{$\sigma_{max}$ in Eqn \ref{eqn:uncertainty_mapping}}        	& $2.0$    &  & \text{$v_{decay}$ in Sec \ref{sec:decay}}        & $5e^{-6}$    \\
\hline
\end{tabular}
\egroup
\label{tab:hyperparameter}
\end{table}


\subsection{Qualitative Comparison on Various Scenarios}
To demonstrate that the network can perform resilient behaviors in different uncertainty environments, we introduce the following three policies in the same navigation network:

\begin{itemize}
    \item \textbf{Over-confident policy}: the navigation network does not have any uncertainty input;
    \item \textbf{Balanced policy}: the navigation network with the uncertainty generated by the predictor, which balances the efficiency and safety;
    \item \textbf{Over-conservative policy}: the navigation network with the double uncertainty generated by the predictor. 
\end{itemize}
In the experiment, we use green, yellow, and red colors to distinguish trajectories generated using the three polices mentioned above. 

\subsubsection{\textit{Encounter}}
To verify whether the expected behavior proposed in \prettyref{sec:network} is learned, we first build the encounter scenario as in \prettyref{fig:f2f_demo}. Note that the behavior of the dynamic obstacle, which goes to the goal without any avoidance action, is unseen in the training scenario. The experimental results are shown in \prettyref{fig:f2f}, where the over-conservative policy can generate a safer trajectory than the balanced policy, which matches our expectation. The over-confident policy behaves too aggressively with respect to the unseen behavior and expects the obstacle to avoid it. Although the robot eventually brakes, it did not avoid being passively hit by the obstacle. 



\subsubsection{\textit{Narrow Corridor}}

\begin{figure}
\centering
\begin{subfigure}{0.23\textwidth}
\centering
\includegraphics[trim={40 0 45 0},clip, width=1\linewidth]{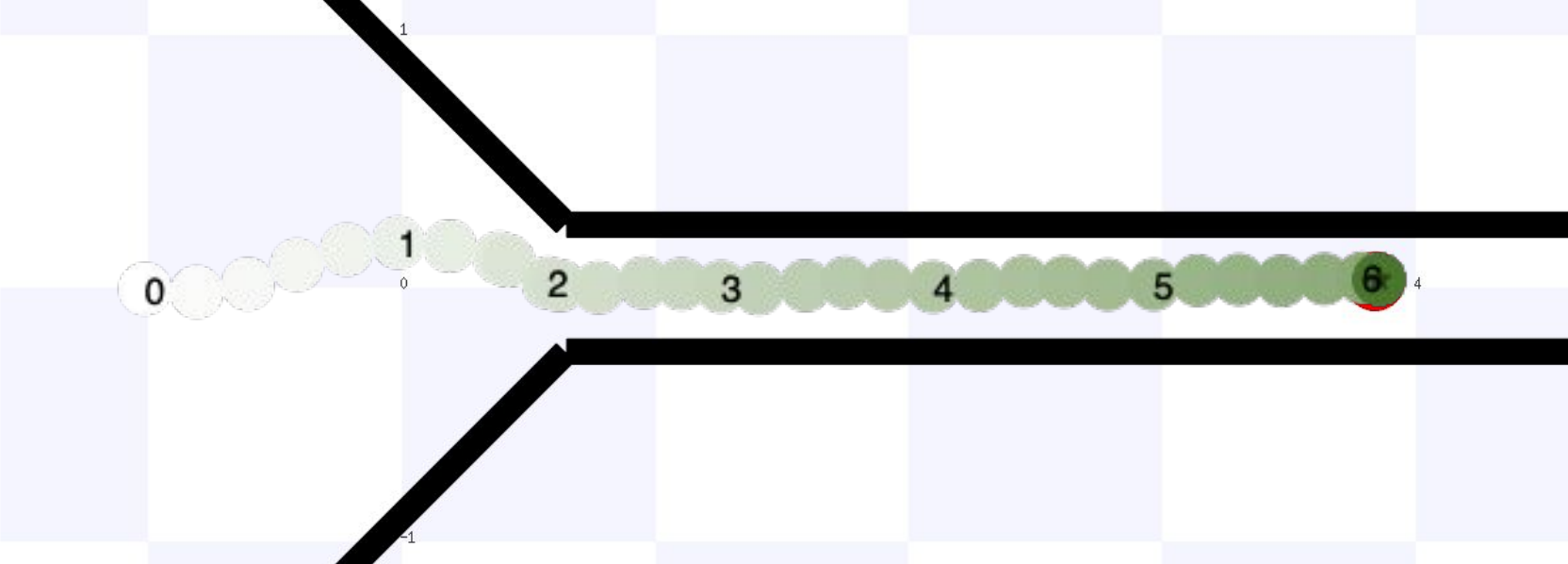}
\caption{Over-confident policy}
\label{fig:narrow_corridor_0}
\end{subfigure} 
\begin{subfigure}{0.23\textwidth}
\centering
\includegraphics[trim={40 0 45 0},clip,width=1\linewidth]{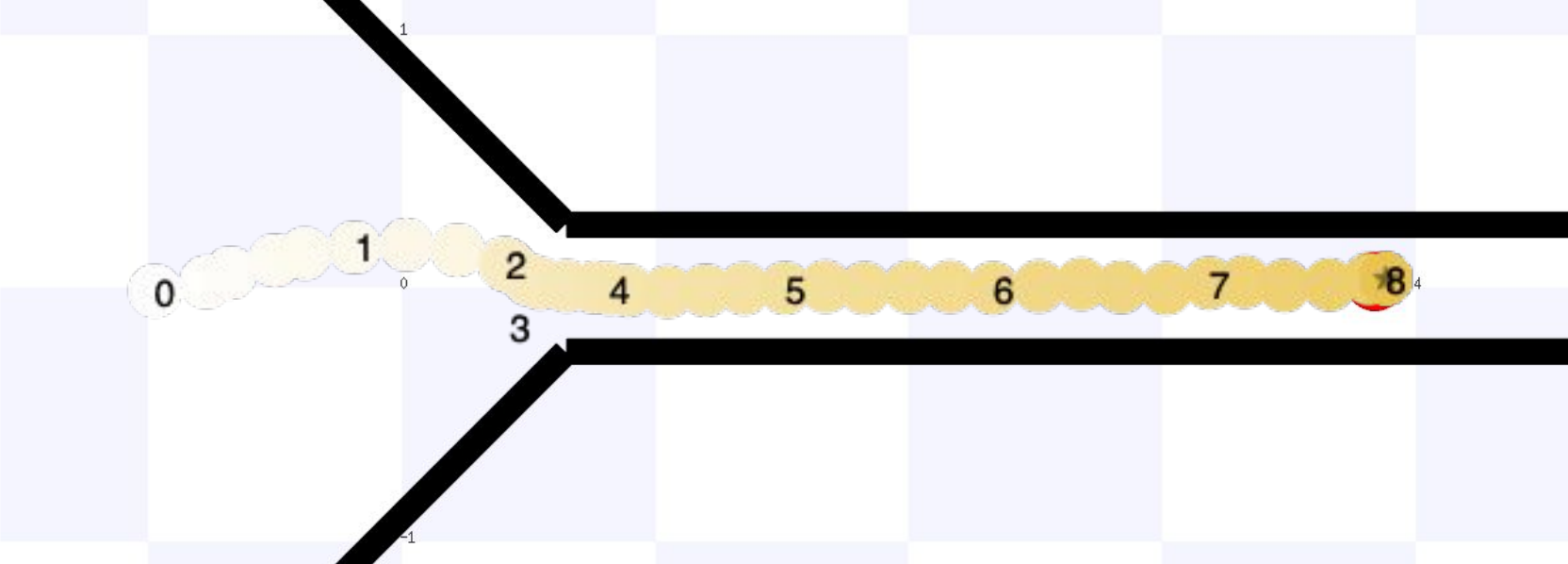}
\caption{Balanced policy}
\label{fig:narrow_corridor_1}
\end{subfigure} \\
\begin{subfigure}{0.23\textwidth}
\centering
\includegraphics[trim={40 0 45 0},clip,width=1\linewidth]{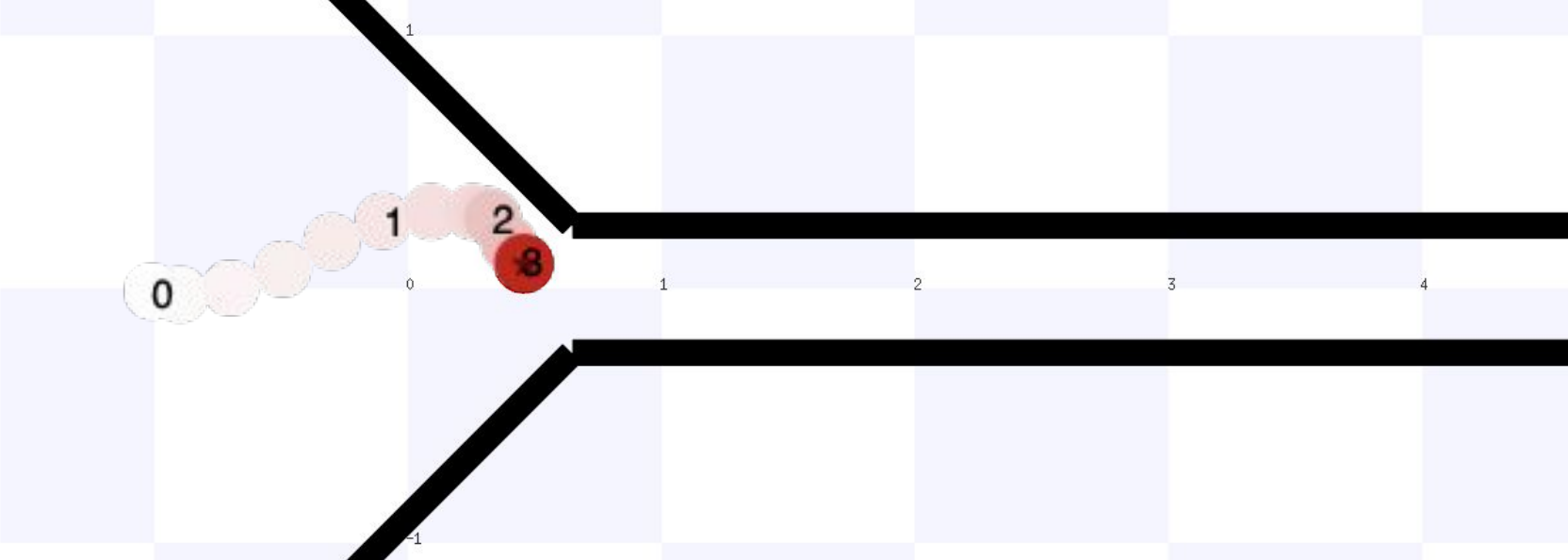}
\caption{Over-conservative policy}
\label{fig:narrow_corridor_2}
\end{subfigure} 
\caption{The experimental results on the narrow corridor scenario. The checkerboard's resolution is $1$ \SI{}{m}. The behavior of the conservative policy is too conservative to enter the narrow corridor. Our balanced policy can make a trade-off between efficiency and safety. }
\label{fig:narrow_corridor}
\vspace*{-0.05in}
\end{figure}

We have demonstrated above that robots can produce more conservative behavior in environments with higher uncertainty. In this experiment, we want to figure out whether the navigation network can perform more aggressive behaviors in the scenario with lower uncertainty. For this purpose, a static, narrow corridor scenario is built. As shown in \prettyref{fig:narrow_corridor}, an over-confident policy is most efficient. After some hesitation at the entrance to the corridor, the balanced policy can also reach the goal without losing much efficiency. For the over-conservative policy, the robot repeatedly hesitates at the entrance and eventually fails to enter the corridor. In this way, we have proved that the navigation network can automatically trade efficiency for safety based on uncertainty in the environment due to the uncertainty mapping function proposed in \prettyref{sec:network}. 

\subsubsection{\textit{Uncertain Corridor}}

\begin{figure}
\centering
\includegraphics[width=1.0\linewidth]{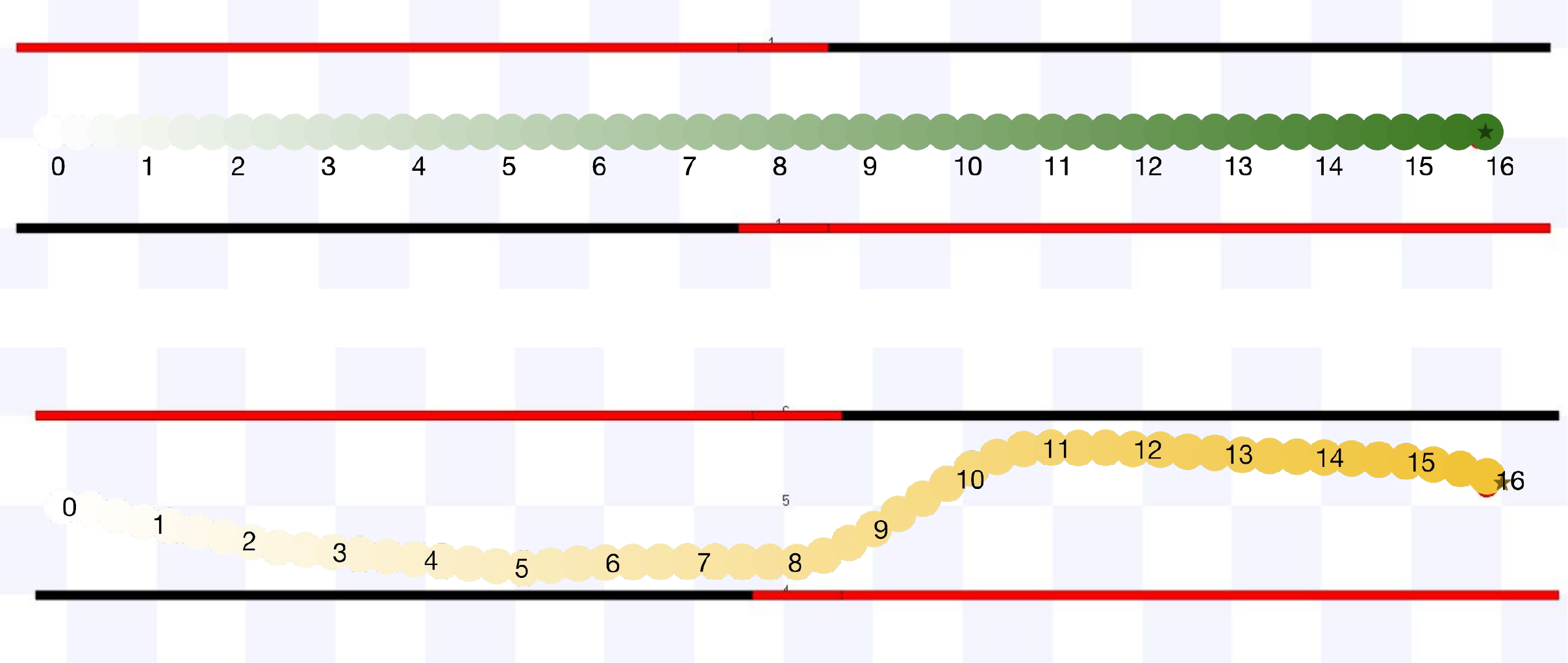}
\caption{The comparison experiments on the uncertain corridor scene. The red obstacles would have a higher uncertainty than the black ones when being perceived by robots. The over-confident policy (green) without uncertainty-awareness would move directly toward the goal. The balanced policy (yellow) will take uncertainty-averse actions to keep away from red obstacles.}
\label{fig:dynamic_corridor}
\vspace*{-0.05in}
\end{figure}

In previous experiments, we have validated that our approach can automatically perform conservative behaviors and aggressive behaviors without any parameter-tuning. In this part, we further investigate whether the network has learned uncertainty-averse behavior. In this experiment, we construct the uncertain corridor scenarios as show in \prettyref{fig:dynamic_corridor}, which have the shape but with different uncertainties. For the over-confident navigation policy, the robot will reach the goal directly. For the balanced navigation policy, however, the robot prefers the area with lower uncertainty to avoid risk although it may be a little inefficient. Hence, uncertainty-averse behavior has been learned by our approach. 

\subsection{Quantitative Experiments with Sensor Malfunction}

To quantitatively evaluate the performance of the uncertainty-aware navigation network, we propose three evaluation metrics: the \textit{success rate}, the \textit{stuck rate}, and the \textit{collision rate}. If the robot can reach its target without any collisions in a limited time, the robot succeeds in the navigation task. If the robot collides or cannot reach the destination within a limited time, they are considered as getting collisions or stuck, respectively. We evaluate the learned uncertainty-aware navigation policy with the uncertainty-unaware baseline in a sensor malfunction scenario. In particular, the scenario includes 20 robots whose initial and goal positions are randomly generated in a $12\times 12$ plane. During the training process, the field of view (FOV) of the 2D LiDAR is $\ang{180}$, and the angular resolution is $\ang{1}$. During testing, we compare our policy to the baseline that does not consider uncertainty. We evaluate all methods on four different sensor malfunction cases where the 2D LiDAR has the effective FOV of $\ang{150}$, $\ang{120}$, $\ang{90}$ and $\ang{60}$ respectively. As we can not change the input dimension of the neural network, we mask the scan part out of the effective FOV to a small fixed value (e.g., 0.1). We evaluate each test with 50 repeats. 

\begin{figure}
\centering
\includegraphics[trim={50 0 85 0},clip, width=0.9\linewidth]{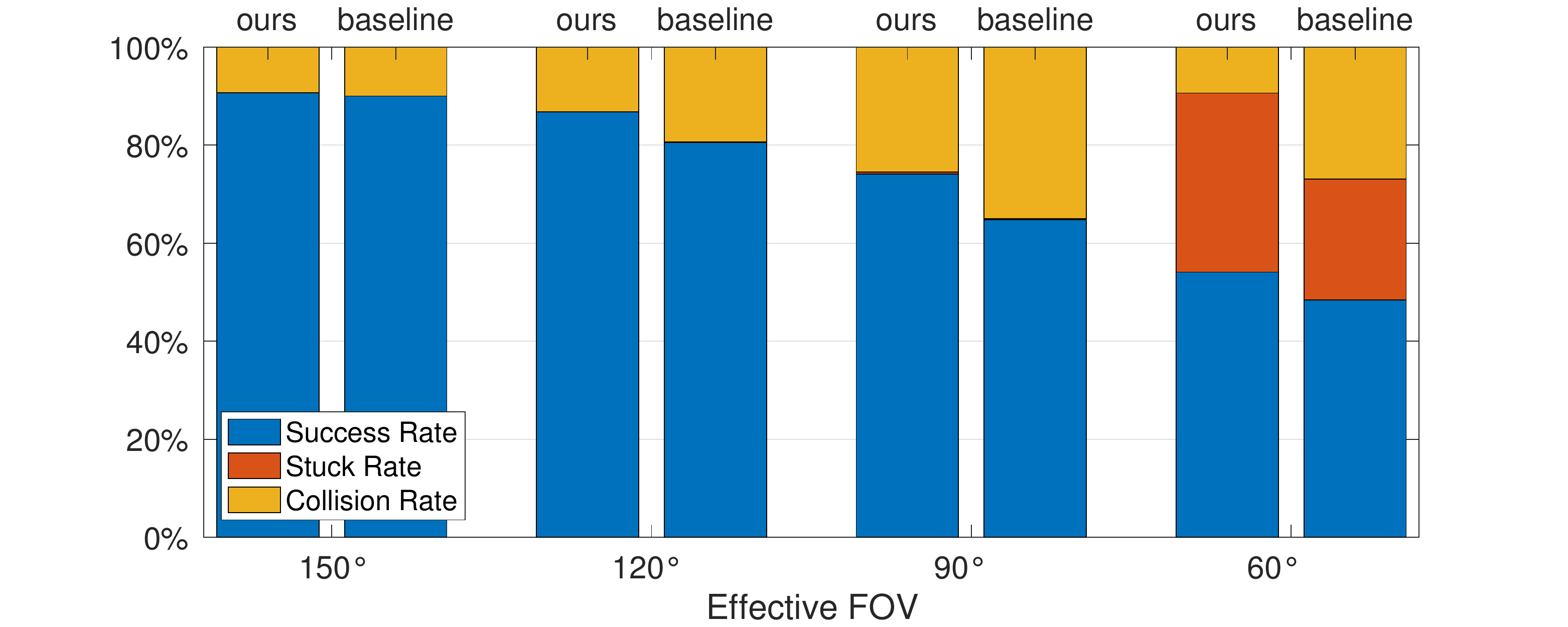}
\caption{Comparison of navigation performance using different effective FOVs.}
\label{fig:fov_results}
\vspace*{-0.05in}
\end{figure}

As shown in \prettyref{fig:fov_results}, our method always outperforms the baseline method in terms of \textit{success rate} and \textit{collision rate}. When the effective FOV decreases, the \textit{success rate} of both methods decreases and the performance gap between them becomes larger. Note that both navigation networks will get stuck in some cases and eventually cannot reach the goal when their FOVs are limited to $\ang{60}$. In this case, the \textit{stuck rate} of our method is higher than the baseline, because our algorithm behaves more cautiously, which from the other side demonstrates that our method can perform safer behaviors in unseen environments. 

%% file: conclusion.tex
\section{Conclusion and Future Work}
\label{sec:conclusion}

This work has developed an model-free uncertainty-aware reinforcement learning to learn resilient navigation behaviors in unseen environments. The core idea is to bridge the gap between the environmental uncertainty and the behavior uncertainty to keep a safe but efficient distance from obstacles and perform uncertainty-averse behaviors. However, our approach strongly relies on the uncertainty estimation but the 2D laser scan is limited to provide an accurate estimation. In future work, our method can be combined with advanced scenario understanding methods to generate more accurate uncertainty based on the semantics of the obstacles. For instance, by setting a high uncertainty to pedestrians, robot navigation can maintain an appropriate distance with pedestrians without intruding their social space.